\documentclass[lettersize,journal]{IEEEtran}
\usepackage{amsmath,amsfonts}
\usepackage{algorithmic}
\usepackage{array}
\usepackage[caption=false,font=normalsize,labelfont=sf,textfont=sf]{subfig}
\usepackage{textcomp}
\usepackage{stfloats}
\usepackage{url}
\usepackage{verbatim}
\usepackage{graphicx}
\usepackage{mathrsfs}
\usepackage{epsfig}
\usepackage{cite}
\usepackage{bbm}
\def\nb0{{\mathbf{0}}}
\def\nb1{{\mathbf{1}}}

\def\ncalA{{\mathcal{A}}}

\def\ncalC{{\mathcal{C}}}
\def\ncalD{{\mathcal{D}}}

\def\ncalJ{{\mathcal{J}}}

\def\ncalN{{\mathcal{N}}}

\def\ncalS{{\mathcal{S}}}

\def\nbbN{{\mathbb{N}}}

\def\nbbR{{\mathbb{R}}}

\newtheorem{defn}{Definition}
\newtheorem{nrem}{Remark}

\newtheorem{assumption}{Assumption}
	
\def\argmin{\operatorname{arg~min}}

\def\E{\mathbb{E}}

\def\snr{\mathtt{SNR}}
\def\jnr{\mathtt{JNR}}

\usepackage[ruled,vlined]{algorithm2e}

\def\by{\bar{y}}
\def\bs{\bar{s}}
\def\bj{\bar{j}}
\def\bn{\bar{n}}

\begin{document}

\title{Linear Jamming Bandits: Sample-Efficient \\Learning for Non-Coherent Digital Jamming}

\author{Charles E. Thornton and R. Michael Buehrer
\thanks{The authors are with Wireless @ Virginia Tech, Bradley Department of ECE, Blacksburg, VA 24061. Correspondence: $thorntonc@vt.edu$.}
}



\maketitle

\begin{abstract}
It has been shown (Amuru \emph{et al.} 2015) that online learning algorithms can be effectively used to select optimal physical layer parameters for jamming against digital modulation schemes without \emph{a priori} knowledge of the victim's transmission strategy. However, this learning problem involves solving a multi-armed bandit problem with a mixed action space that can grow very large. As a result, convergence to the optimal jamming strategy can be slow, especially when the victim and jammer's symbols are not perfectly synchronized. In this work, we remedy the sample efficiency issues by introducing a linear bandit algorithm that accounts for inherent similarities between actions. Further, we propose context features which are well-suited for the statistical features of the non-coherent jamming problem and demonstrate significantly improved convergence behavior compared to the prior art. Additionally, we show how prior knowledge about the victim's transmissions can be seamlessly integrated into the learning framework. We finally discuss limitations in the asymptotic regime.
\end{abstract}


\section{Introduction}
Broadly speaking, a radio jamming system intentionally transmits energy to disrupt reliable data communication \cite{amuru2015optimal,amuru2015jamming,Lichtman2016}. Evaluating the impact of adversarial jamming attacks is a crucial concern in the design wireless protocol \cite{lichtman20185g}. Further, effective strategies for communications denial are of marked interest for military applications where untrusted communication must be stopped.

Jamming techniques have historically been studied in the context of spread spectrum communications, but more recently interest has grown in evaluating the susceptibility of a broader class of systems to jamming attacks. Many works have analyzed the performance of jamming via optimization or information theoretic techniques \cite{mceliece1981information,Basar1983}. Unfortunately, many of these analytical works assume the jammer has access to information about the victim's transmission strategy or the channel, which is rarely the case in adversarial scenarios.

With the rise of adaptive transmission strategies, especially in the context of machine-learning enabled communication systems, it is particularly important to understand the statistical behavior of optimal jamming systems. Work in this direction has been carried out in \cite{amuru2015optimal,amuru2015jamming}. In \cite{amuru2015optimal}, the optimal physical layer jamming strategies are analytically derived for several cases of interest. However, implementation of these optimal strategies requires \emph{a priori} knowledge about the victim's transmission strategy. In \cite{amuru2015jamming}, a multi-armed bandit (MAB) learning algorithm is presented that is guaranteed to converge to the optimal jamming strategy without making any assumptions about the victim's transmissions. However, this convergence can be slow as the jammer's strategy set grows, and is especially cumbersome in the case of non-idealities such as phase or timing offset. For realistic jamming scenarios, the action space is likely to grow so large that it is highly impractical to maintain a separate confidence bound for each jamming strategy, as employed in \cite{amuru2015jamming}. In this paper, we focus on the non-coherent jamming scenario, and present a practical learning algorithm that accounts for the inherent similarity between jamming strategies and the unique statistical features of the non-coherent scenario. 

\emph{Contributions:} Herein, we attempt to characterize the finite-time behavior of an intelligent jamming system which exploits inherent similarity between transmission strategies to reduce convergence time. Although the jamming bandits algorithm is guaranteed to converge to the optimal strategy in the limit, electronic-warfare scenarios are often time-sensitive. Thus, it is expected that improved convergence rates will be of more practical value than asymptotic guarantees. In particular, we examine the difficult case of non-coherent jamming, in which the effectiveness of the jamming signal is impeded by an unknown phase offset, modeled as a uniform random variable. We show that this case introduces a heavy-tailed distribution for the optimal jamming strategy, and design a linear contextual bandit algorithm that uses appropriate context features for efficient non-coherent jamming. We show that this algorithm results in approximately an order of magnitude improvement in convergence rate over the UCB-1 algorithm proposed in \cite{amuru2015jamming}. Further, the linear contextual bandit algorithm allows the jammer to scale to very large strategy spaces with minimal computational burden. We further show how prior information about the victim's transmission strategy can be seamlessly integrated into the learning framework by incorporating additional context features.

\section{System Model}
\label{se:system}
In this section, we review several known results related to optimal jamming against digital modulation schemes \cite{amuru2015optimal}. These modeling assumptions clarify the scenario of interest and shed light on important considerations for the linear bandit learning problem discussed in Sections \ref{se:learning} and \ref{se:linear}.

It is assumed that the data transmitted by the victim is mapped onto a known digital amplitude phase constellation. The low-pass representation of the transmitted signal is given by 
\begin{equation}
	s(t) = \sum_{m=-\infty}^{\infty} \sqrt{P_{S}}s_{m}g(t-mT),
\end{equation}
where $P_{S}$ is the average received power, $g(t)$ is the real-valued pulse shape, $s_{m}$ are the modulated symbols, and $T$ is the symbol interval. Each symbol is transmitted with uniform probability. Without loss of generality, we assume $\E[|g(t)|^{2}] = \E[|s_{m}|^{2}] = 1$. The victim's transmitted signal passes through and AWGN channel and is corrupted by a jamming signal, given by
\begin{equation}
	j(t) = \sum_{m=-\infty}^{\infty} \sqrt{P_{J}}j_{m}g(t-mT),
\end{equation}
where $P_{J}$ is the average jamming signal power as seen at the victim's receiver and $j_{m}$ are the modulated jamming symbols for which $\E[|j|^{2}] = 1$.

From the jammer's perspective, the ideal case occurs when the jamming signal is both phase and time synchronous with the victim's signal. Although we are presently interested in the non-coherent case, we will briefly discuss the coherent case for expository purposes. After matched filtering and sampling once per symbol interval, the received signal at the victim is given by
\begin{equation}
	y_{k} = y(t=kT) = \sqrt{P_{s}}s_{k}+\sqrt{P_{J}}j_{k} + n_{k},
\end{equation}
where $k = 1,2,...$, and $n_{k}$ is a zero-mean Gaussian noise term with variance $\sigma^{2}$. The victim's signaling scheme, jammer's signaling scheme, and noise terms are assumed to be statistically independent. We may define $\snr = \frac{P_{S}}{\sigma^{2}}$ and $\jnr = \frac{P_{J}}{\sigma^{2}}$ to be the signal-to-noise and jammer-to-noise ratios, respectively. For 2-dimensional constellations, such as $M$-QAM, we introduce the notation $\bar{y}_{k} = [\Re_{y_{k}},\Im_{y_{k}}]$, where $\Re_{y_{k}}$ and $\Im_{y_{k}}$ correspond to the real and imaginary parts of $y_{k}$, respectively. Similarly defining $\bar{s}_{k}$, $\bar{j}_{k}$, and $\bar{n}_{k}$, the received signal may be expressed as
\begin{equation}
	\bar{y}_{k} = \sqrt{P_{S}}\bar{s}_{k}+\sqrt{P_{J}}\bar{j}_{k}+\bar{n}_{k}.
\end{equation}

\begin{figure}
	\centering
	\includegraphics[scale=0.6]{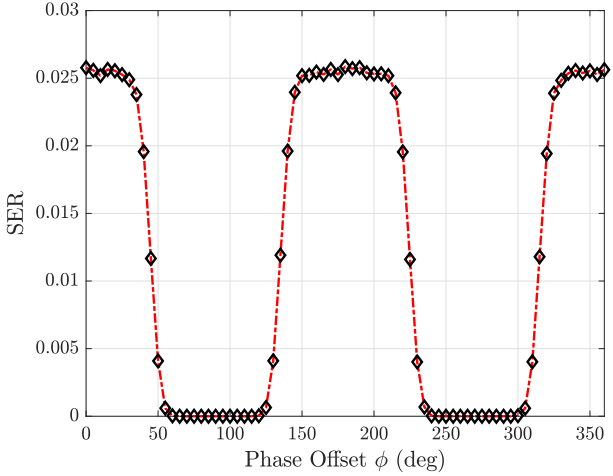}
	\caption{\textsc{Phase offset vs SER} for the optimal jamming strategy against a victim using BPSK at $\jnr = 10$ dB and $\snr = 20$ dB. When the symbols are approximately $90$ degrees out-of-phase, the SER is near zero, even if the optimal jamming strategy is used. This presents a challenge for online learning algorithms, since optimal signaling strategies may yield a very high cost depending on the phase offset.}
	\label{fig:offset}
\end{figure}

The average probability of error at the victim's receiver can be viewed as two orthogonal $\sqrt{M}$-PAM signals and is expressed by
\begin{multline}
	p_{e}(\bj,\snr,\jnr) \approx \\ \left( 1-\frac{1}{\sqrt{M}} \right) \frac{1}{2} \times \bigg[ \operatorname{erfc} \left(\sqrt{\snr}\frac{d_{\text{min}}}{2}+\sqrt{\jnr}j \right) \\ +\operatorname{erfc} \left(\sqrt{\snr}\frac{d_{\text{min}}}{2}-\sqrt{\jnr}j \right) \bigg],
\end{multline}
where $j = \Re \bar{j}$ or $\Im \bar{j}$, $M$ is the order of the constellation, and $d_{\text{min}}$ is the minimum distance of the constellation. 

In this paper, however, we are primarily concerned with the more realistic non-coherent case, in which there is a random unknown phase offset between the jamming and victim signals. In the non-coherent case, the jammer wishes to utilize a transmission scheme that causes a large probability of error across all possible phase offsets. In this case, the received signal is given by
\begin{equation}
	\by_{k} = \sqrt{P_{S}} \bs_{k} + \sqrt{P_{J}} \operatorname{exp}(i \phi) \bj_{k} + \bn_{k},
\end{equation}
where $i = \sqrt{-1}$ and $\phi \sim \mathrm{unif}(0,2\pi)$ is the random phase offset. In the non-coherent case, the probability of error along the in-phase direction can similarly be written as
\begin{multline}
		p_{e}(\bar{j}, \snr, \jnr) \approx 
		\left(1-\frac{1}{\sqrt{M}}\right) \frac{1}{2} \\ \times \bigg[ \operatorname{erfc}\left(\sqrt{\operatorname{\snr}} \frac{d_{\min }}{2} +\sqrt{\mathrm{\jnr}}(\Re \bar{j} \cos (\phi)-\Im \bar{j} \sin (\phi))\right) \\
		+\operatorname{erfc}\left(\sqrt{\operatorname{\snr}} \frac{d_{\min }}{2}-\sqrt{\mathrm{\jnr}}(\Re \bar{j} \cos (\phi)-\Im \bar{j} \sin (\phi))\right) \bigg].
\end{multline}

A similar expression holds true for the quadrature dimension. In order to maximize the probability of error for fixed values of $P_{S}$ and $P_{J}$, the following optimization problem can be solved
\begin{equation}
	\underset{f_{\bj}}{\max} \; \E_{f_{\bj}} \left[\E_{\phi}[p_{e}(\bj,P_{S},P_{J})] \right] \quad \mathrm{s.t.} \quad \E[\lVert \bj \rVert^{2}] \leq 1
\end{equation}
which is numerically solved in \cite{amuru2015optimal} for several cases of interest. It is found that the optimal jamming signal does not have a fixed power level, but instead has a pulsed structure. The optimal jammer transmits power level $\jnr/\rho$ with probability $\rho$, and at power level $0$ with probability $1-\rho$.   

Unfortunately, in a realistic communications scenario, information about the victim's transmission strategy is likely unknown \emph{a priori} and can be cumbersome to obtain. Thus, a practical jamming system may be expected to learn the optimal jamming strategy using real-time feedback. This involves the classic problem of exploration and exploitation that was investigated in \cite{amuru2015jamming}. 

However, the development in \cite{amuru2015jamming} is primarily concerned with the asymptotic optimality of bandit learning, given that the UCB-1 algorithm is used. As a result, no guarantees are made as to the finite-time behavior, which is seen to be especially poor in the non-coherent jamming scenario when UCB-1 is used \cite[Chapter 7]{Lattimore2020bandit}. In the following discussion, we introduce a linear bandit approach to real-time jamming which significantly improves finite-time behavior by making use of inherent similarity between jamming strategies and specific statistical features which are relevant for the non-coherent jamming problem.

\begin{figure*}[t]
	\centering
	\includegraphics[scale=0.56]{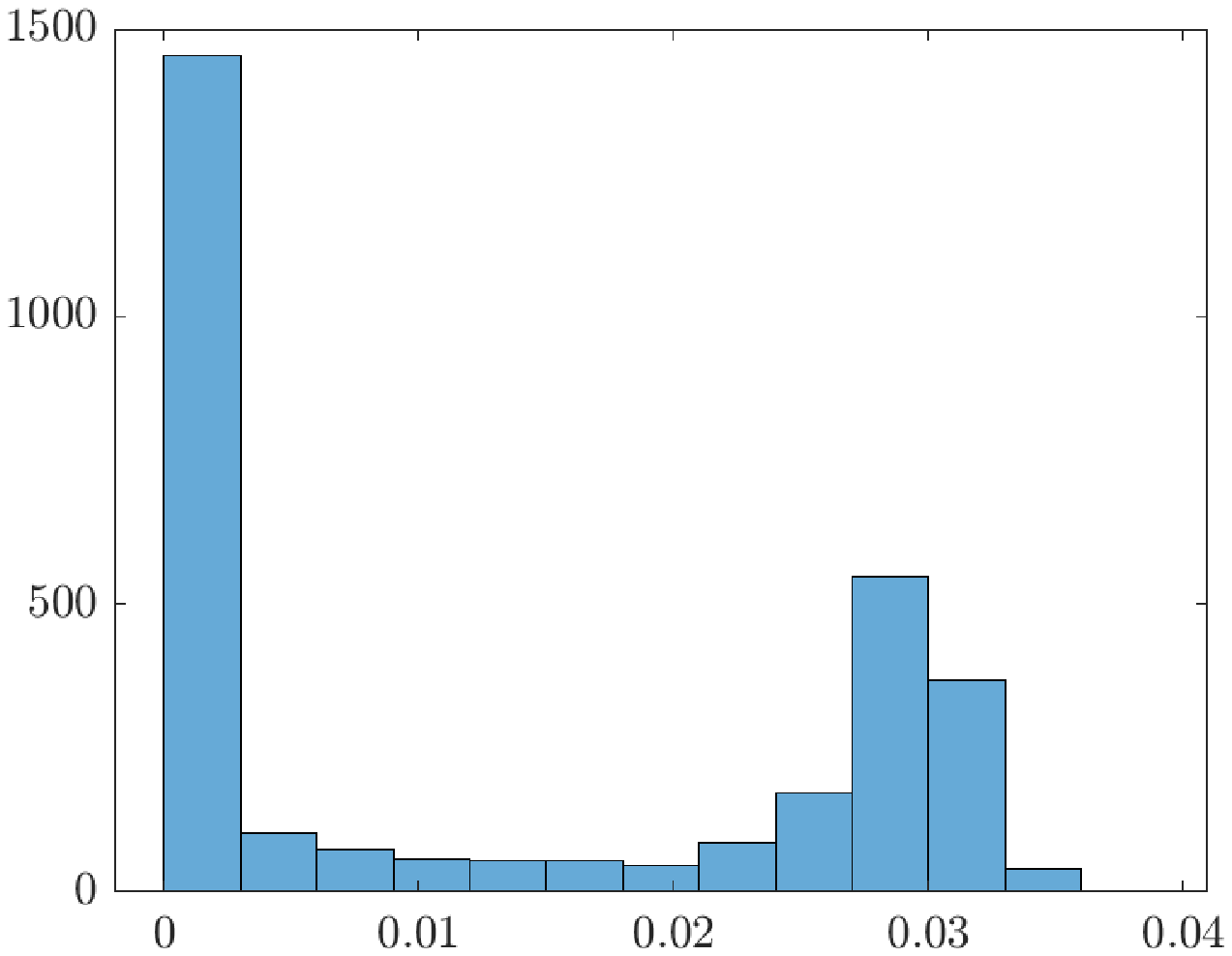}
	\includegraphics[scale=0.56]{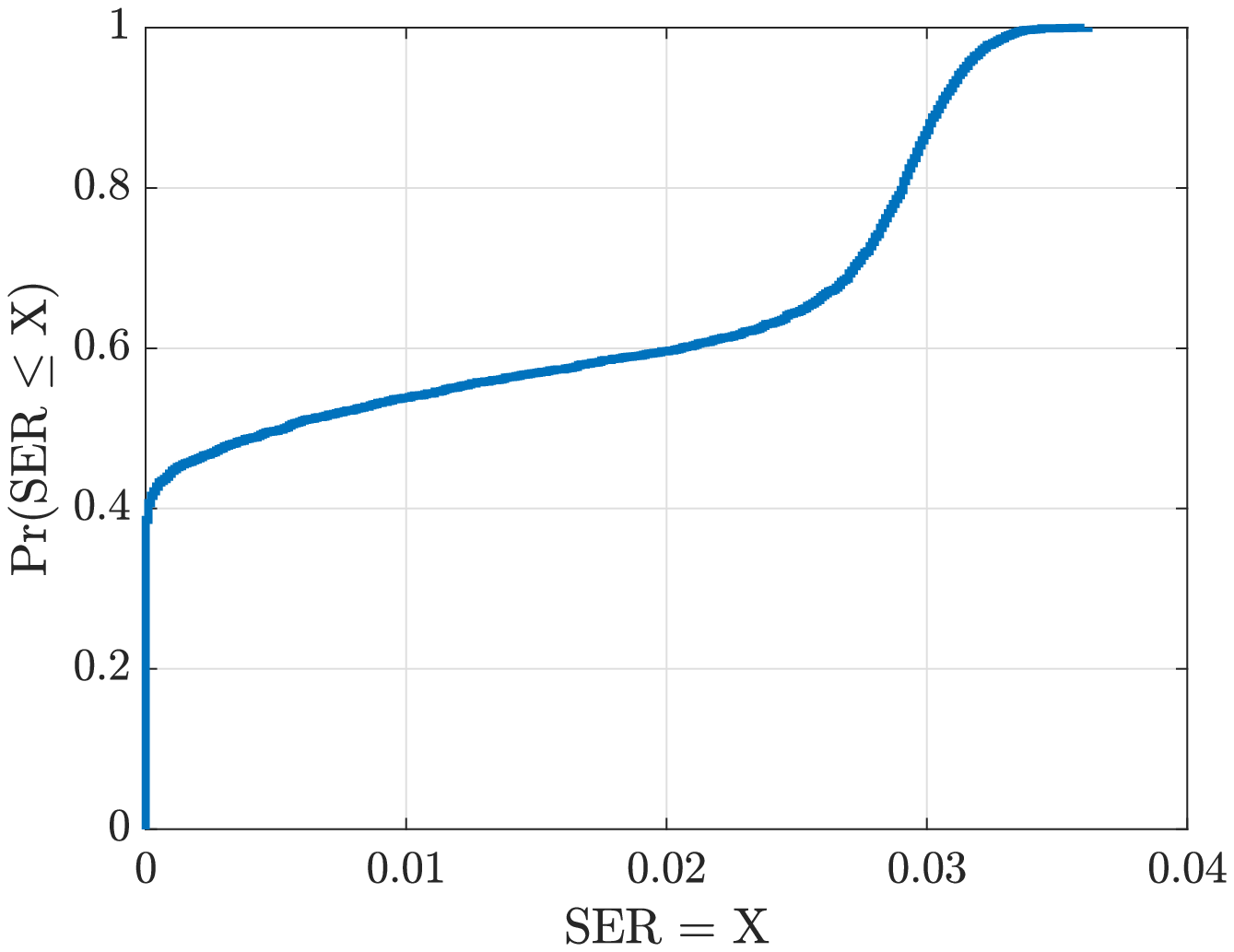}
	\caption{Distribution of $\mathrm{SER}$ for the optimal jamming strategy in non-coherent jamming scenario. Even when the optimal strategy is used, a SER of zero is frequently observed. Thus, more nuanced features than the expected cost of each arm are desirable for learning in the non-coherent case so that near-optimal strategies are not eliminated too quickly. Our choice of context features are seen in (\ref{eq:context}).}
	\label{fig:dist}
\end{figure*}

\section{Learning Problem}
\label{se:learning}
At each time step $t$, the jammer chooses an action $a_{t} \in \ncalA$, which is composed of three components: $a_{t} = [\text{Signaling scheme},\jnr,\rho]$. The choice of signaling scheme is made from set $\ncalJ = \{\text{BPSK},\text{QPSK},\text{AWGN}\}$. The choice of $\jnr$ and $\rho$ are made from a continuous strategy set $\ncalS$, composing values of $\jnr \in [\jnr_{min},\jnr_{max}]$ and $\rho \in [0,1]$, is a compact subset of $(\nbbR_{+})^{2}$. The cardinality of the action set $\ncalA$ depends heavily on how the continuous parameters $\jnr$ and $\rho$ are discretized. Here, we introduce a discretization parameter $M$, and discretize $\rho$ as $\{1/M,2/M,...,1\}$ and $\jnr$ as $\jnr_{\text{min}}+(\jnr_{\text{max}}-\jnr_{\text{min}}) \times \{1/M,2/M,...,1\}$. In \cite{amuru2015jamming}, a procedure for learning an optimal discretization parameter is proposed, using specific features of the cost function. Here, we do not treat such a problem and focus on improving the speed of learning for a given discretization factor. However, we will examine learning performance for several discretization factors in Section \ref{se:sim}.

After an action is selected, the jammer receives a real-valued cost $C_{t}: \{\ncalJ, \ncalS \} \mapsto \nbbR$. The exact structure of the cost function is unknown to the jammer \emph{a priori} and must be learned through repeated experience.

\begin{defn}[SER Cost Function]
	Let $\jnr_{t}$ be the average $\jnr$ used by the jammer at time $t$. Let $\mathrm{SER}_{t}$ be the symbol error rate observed at the victim when the jammer uses a particular strategy $\{j_{k} \in \ncalJ, s_{k} \in \ncalS \}$ and let $\mathrm{SER}_{\mathrm{target}}$ be a target symbol error rate. An effective cost function is then
	\begin{equation}
		C_{t} = \max(\mathrm{SER}_{t}-\mathrm{SER}_{target},0)/\jnr_{t}
	\end{equation}
\end{defn}

\begin{defn}[PER Cost Function]
	Using similar notation, a more realistic cost function can be defined in terms of packet-error rate as
	\begin{equation}
		C_{t} = \max(\mathrm{PER}_{t}-\mathrm{PER}_{target},0)/\jnr_{t}
	\end{equation}
\end{defn}

Both of the above cost functions satisfy the $\alpha$-Hölder condition, meaning that for any set of strategies used by the victim and jammer, the cost function is locally Hölder continuous. This implies that jamming strategies with similar parameters will result in similar expected costs. In \cite{amuru2015jamming}, the $\alpha$-Hölder condition was used to prove asymptotic near-optimality of the UCB-1 algorithm for learning the optimal jamming strategy. Unfortunately, the UCB-1 algorithm requires that a separate confidence radius be maintained for each action. Thus, as the cardinality of the action space grows, the finite-time behavior of the UCB-1 algorithm becomes infeasible for practical applications. Further, UCB-1 only accounts for the average cost associated with each action. As seen in Figure \ref{fig:offset}, this is not sufficient for the non-coherent jamming scenario. In the next section, we propose an algorithm to remedy these concerns. 

\section{Linear Jamming Bandits Algorithm}
\label{se:linear}
An important generalization of the multi-armed bandit problem is called contextual bandits. In the contextual bandit setting, the decision maker is able to utilize side information to improve decision-making capabilities, which often drastically reduces the need for exploration. However, estimating the average cost for each context-action pair can be cumbersome in practice, especially as the context space grows. A salient case of the contextual bandit problem, called linear bandits, assumes that the decision maker can learn about one context by experiencing a similar context \cite{Lattimore2020bandit,abeille2017linear}.

In this section, we describe a practical algorithm in which the context features are statistics about the error performance of a particular jamming strategy. Since the proposed cost functions are $\alpha$-Holder continuous over the action space \cite[Theorem 1]{amuru2015jamming}, we observe that knowledge does transfer particularly well between contexts, given a reasonable choice of context features. The linear jamming bandit problem proceeds as follows. During each step, the jammer makes use of the following context features, defined for each $a_{i} \in \ncalA$:
\begin{multline}
	\varphi_{i}(t=n) = \\ \bigg[ \frac{1}{n}\sum_{t=1}^{n} C_{t}(a_{i}), \frac{1}{n} \sum_{t=1}^{n} \mathbbm{1}\{C_{t}(a_{i}) > \tau\},  \max_{t \leq n} C_{t}(a_{i}) \bigg],
	\label{eq:context}
\end{multline} 
where $\mathbbm{1}\{\}$ is the indicator function, which returns zero if the argument is false and one if the argument is true, and $\tau > 0$ is a threshold selected to indicate successful disruption of the victim's communication. We note that the specific choice of $\tau$ is not overly important, as the objective of the second context feature is to capture the frequency with which a given jamming strategy produces a nonzero error rate.

These context features capture some unique aspects of the non-coherent jamming problem. As seen in Figure \ref{fig:offset}, when the victim transmits BPSK and the jamming and victim signals are close to $90$ degrees out-of-phase, the SER will be close to zero, even if the optimal jamming strategy is used. Thus, if a learning algorithm only considers the expected cost of each arm, the algorithm will be generally insufficient for decision making. 

In (\ref{eq:context}), the latter two features capture more relevant information that aids in discerning the utility of jamming strategies, since it is assumed that very low error rates will be experienced frequently due to the unknown phase offset. This intuition is further confirmed by Figure \ref{fig:dist}, which shows the distribution of SER for the optimal jamming strategy when the victim transmits BPSK. We note that the distribution for the optimal jamming strategy is bimodal, with one mode occurring at $\mathrm{SER} = 0$ and another mode occurring at $\mathrm{SER} \approx 0.03$, corresponding to the case where the signals are mostly in-phase. Since both optimal and highly suboptimal jamming strategies will frequently result in error rates of zero, the use of additional context features is crucial for the non-coherent jamming problem.

\begin{nrem}
The context features listed in (\ref{eq:context}) are by no means unique, but were found to be effective for the cases tested in simulation. Additionally, these context features do not assume any access to additional information about the victim's symbol.
\end{nrem}

The jammer's goal is then to learn a weighting vector $\theta \in \nbbR^{3}$ which predicts the cost for each action and allows the jammer to select a reasonable strategy. This is made possible via the following `linearity' assumption:
\begin{assumption}[Stochastic Linear Bandit Structure]
	Let $\{\varphi_{i}\}$ be a set of context features defined for each $a_{i} \in \ncalA$ and let $\theta$ be a weighting vector. Then the following relationship holds for all $a_{i} \in \ncalA$ and for all time $t \in \nbbN_{+}$
	\begin{equation}
		C_{t}(a_{i}) = \langle \varphi_{i} , \theta \rangle + \eta_{t},
	\end{equation}
	where $\eta_{t}$ is a conditionally sub-Gaussian random variable, conditioned on the jammer's knowledge of the history of costs, actions, and contexts.
	\label{assum:lin}
\end{assumption}

\begin{nrem}
	The linear bandit framework is reasonable when the context features $\{\varphi_{i}\}_{i \in \ncalA}$ can be used to accurate differentiate the value of each arm. This implies that the context set is ``rich" enough to describe the statistics of the cost distribution for each arm. From the simulations in Section \ref{se:sim}, we observe that Assumption \ref{assum:lin} is indeed pragmatic for the non-coherent jamming problem given the context features defined in (\ref{eq:context}).
\end{nrem}

\begin{algorithm}[t]
	\setlength{\textfloatsep}{0pt}
	\caption{Linear Jamming Bandits}
	\SetAlgoLined
	\textbf{Input} Discretization Factor $M$, Cost function $\ncalC$, $B = I_{d}$, $\hat{\mu} = 0_{d}$, $f = 0_{d}$\\
	\For{\text{Each time step $t= \; 1,...,T$}}{
		\vspace{0.07cm}
		(1) Sample $\tilde{\mu}_{t} \sim \ncalN(\hat{\mu},B^{-1})$;\\ \vspace{.2cm}
		(2) Assemble context vectors using (\ref{eq:context});\\ \vspace{0.2cm}
		(3) Utilize jamming strategy $a_{t} = \argmin_{i} \langle \varphi_{i}, \tilde{\mu}_{t} \rangle$ and observe cost $\ncalC_{t}$; \\ \vspace{.2cm}
		(4) Update $B = B + \varphi_{a_{t}} \varphi_{a_{t}}^{T}$, $f = f + \varphi_{a_{t}} \ncalC_{t}$, and $\hat{\mu} = B^{-1}f$.
		\label{algo:linjb}
	}
\end{algorithm}

To solve the linear bandit problem, we utilize a Bayesian-inspired approach called Thompson Sampling (TS) which is well-known to provide good theoretical and empirical performance alike in the linear bandit setting \cite{agrawal2013thompson,abeille2017linear}. The basic idea of TS is to select actions based on the posterior probability that they will provide the highest reward. This is efficiently performed in practice by assuming a conjugacy relation between the prior and posterior distributions of the parameter vector $\theta$. Once a posterior distribution $P(\theta|\ncalD)$ is estimated, samples can be easily obtained and actions are selected which minimize the inner product relationship $\E[C(a_{i})] = \langle \varphi_{i}, \theta \rangle$. 

The linear TS algorithm used to obtain the results found in this paper is seen in Algorithm \ref{algo:linjb}. To obtain a posterior distribution, we set a normally distributed prior on the parameter\footnote{It is important to note that the normal-normal conjugacy assumption need not hold for any physical system parameters, but is simply used to express uncertainty about $\theta$.} $\theta \sim \ncalN(\hat{\theta},B^{-1})$, where $B$ and $\hat{\theta}$ are initialized to uninformative values as shown in Algorithm \ref{algo:linjb}. The posterior parameters are sequentially computed via standard Bayesian updates, for which the detailed computations can be found in \cite[Appendix A.1]{agrawal2013thompson}.

The exploration in the linear TS algorithm comes from the randomization. If the posterior mass is not concentrated around a particular value of $\theta$, the fluctuations in the samples are expected to be large and the jammer will select a variety of actions as a result. As the posterior mass becomes more concentrated with experience, the jammer will settle on a particular action. The advantages of Algorithm 1 as compared to the UCB-1 algorithm proposed in \cite{amuru2015jamming} are as follows:
\begin{enumerate}
	\item A separate confidence radius for each action need not be maintained. This results in faster convergence and better scalability to large action sets.
	\item The context vectors allow for problem-specific features to be captured by the learning algorithm, reducing the need for exploration.
	\item The computation of Algorithm \ref{algo:linjb} is straightforward and limited only by the dimensionality of the context vector.
\end{enumerate}
Unfortunately, the linear TS approach does come with one particular downside: asymptotic convergence to the optimal jamming strategy is not guaranteed. Since the algorithm is randomized, initialization and the particular choice of prior prevent such asymptotic guarantees as are available for UCB-1. However, in the following section, we demonstrate linear-TS significantly improves finite-time behavior in several jamming scenarios.

\section{Simulations}
\label{se:sim}

\begin{figure}[t]
	\centering
	\includegraphics[scale=0.6]{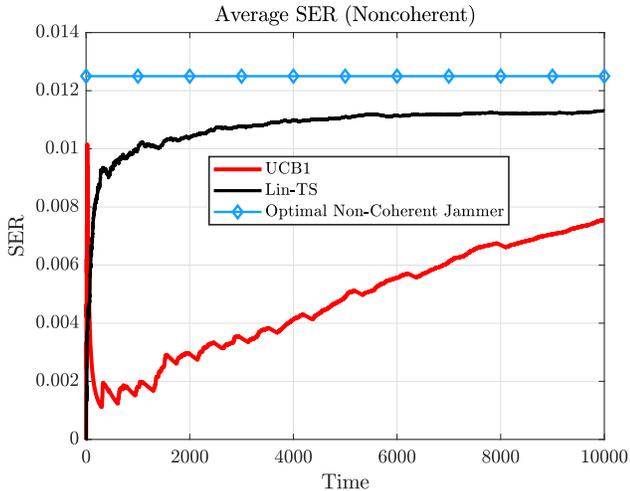}
	\caption{\textsc{Convergence behavior} of proposed linear Thompson Sampling algorithm when jamming a single stationary user. The victim uses BPSK at $\snr = 20$dB. The $\jnr$ is fixed at $10$dB. A discretization factor of $M = 100$ is applied.}
	\label{fig:convBpsk}
\end{figure}

\begin{figure}[t]
	\centering
	\includegraphics[scale=0.6]{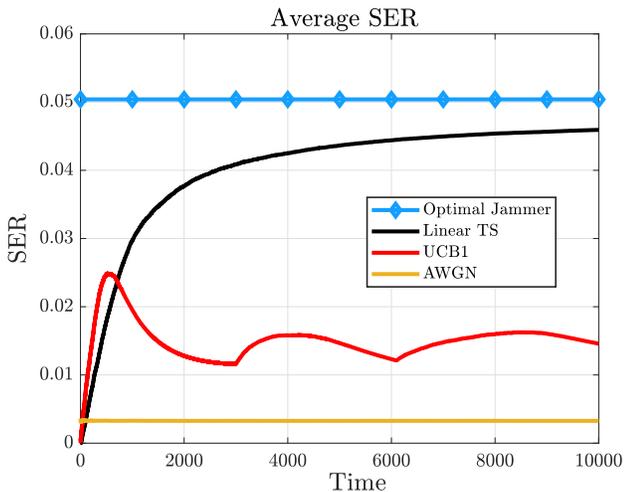}
	\caption{\textsc{Convergence behavior} of proposed linear Thompson Sampling algorithm when jamming a single stationary user. The victim uses QPSK at $\snr = 20$dB. The $\jnr$ is fixed at $10$dB. A discretization factor of $M = 1000$ is applied. The UCB-1 algorithm remains in the exploration phase after $10,000$ packets.}
	\label{fig:convQpsk}
\end{figure}

\begin{figure}
	\centering
	\includegraphics[scale=0.6]{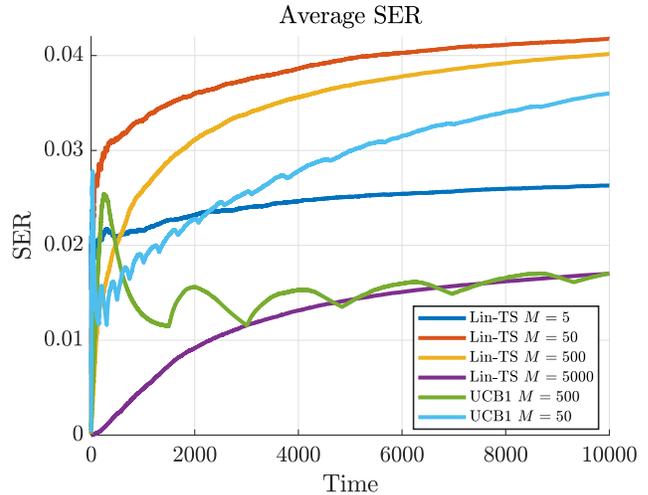}
	\caption{\textsc{Impact of discretization} on linear-TS algorithm compared to UCB-1. We observe smoother convergence behavior when the linear bandit algorithm is employed, and graceful decay as the discretization factor is made very large.}
	\label{fig:disc}
\end{figure}

\begin{figure}
	\centering
	\includegraphics[scale=0.6]{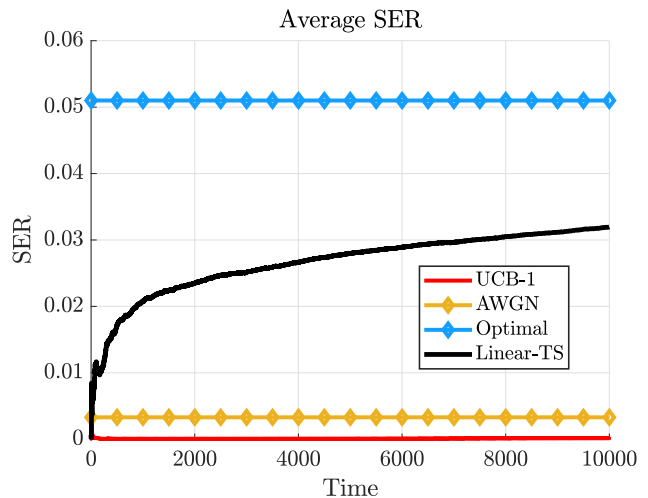}
	\caption{\textsc{Convergence behavior} when $\jnr$ is variable, and the size of the action space increases. The victim uses QPSK at $\snr = 20$dB. A discretization factor of $M=100$ is applied. The UCB-1 algorithm is not able to explore each action once during the finite time horizon.}
	\label{fig:varyJNR}
\end{figure}

\begin{figure}
	\centering
	\includegraphics[scale=0.6]{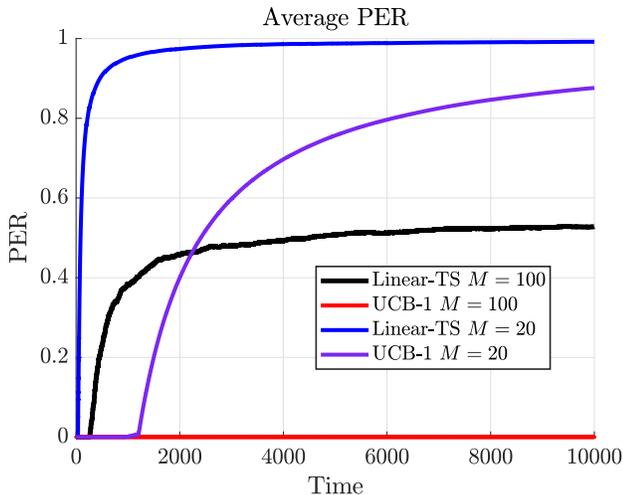}
	\caption{\textsc{Convergence behavior} when the $\mathrm{PER}$ cost function is used. We observe that for both discretization factors, $M=20$ and $M=100$, linear-TS provides a marked performance improvement over UCB-1. In the case of $M=100$, UCB-1 is unable to explore the action space and obtains an average $\mathrm{PER}$ of zero.}
	\label{fig:per}
\end{figure}

In the following simulations, both the jammer and the victim transmit $1$ packet per time step. Each packet consists of $10000$ symbols. The jammer's set of signaling schemes is $\ncalJ = \{\text{AWGN},\text{BPSK},\text{QPSK}\}$ and the victim's signaling set is $\{\text{BPSK},\text{QPSK}\}$. We are interested in characterizing the small-sample behavior of the proposed algorithm, and limit the simulation length to $10000$ time steps.

In Figure \ref{fig:convBpsk}, we observe the convergence behavior of the proposed linear TS algorithm compared to the UCB-1 algorithm proposed in \cite{amuru2015jamming} and the optimal jamming strategy derived in \cite{amuru2015optimal}. The victim uses a fixed transmission strategy of BPSK, and the $\jnr$ is fixed at $10$ dB. The discretization factor $M = 100$. We observe that even in this case, where the action space is small, the proposed linear-TS algorithm still performs much better than the UCB-1 algorithm.

In Figure \ref{fig:convQpsk}, we analyze a similar scenario, except the user employs QPSK and the discretization factor is increased to $M = 1000$. In this case, the cardinality of the action space $|\ncalA| = 3000$ and the burden of exploration placed on the UCB-1 algorithm is much greater, since UCB-1 must maintain a separate confidence radius for each arm, and is guaranteed to play each arm infinitely often in the limiting case. Although the regret bound for UCB-1 can be shown to be asymptotically optimal \cite{Lattimore2020bandit,amuru2015jamming}, this presents a major limitation in the finite-sample regime. We observe that the linear-TS algorithm performs significantly better, achieving an average $\mathrm{SER}$ of $>.045$ in the limited time horizon.

In Figure \ref{fig:disc}, we examine in more detail the impact of the discretization parameter $M$, and therefore the cardinality of the action set $|\ncalA|$ on the convergence behavior of the UCB-1 and linear-TS algorithms. We observe that for a discretization factor $M=5$, the action set is so small that the jammer learns a significantly sub-optimal strategy regardless of the learning algorithm employed. When the discretization factor is increased to $M=50$, the gap in convergence rate between UCB-1 and linear-TS is appreciable, although UCB-1 is still able to explore the entire action space in the finite window and convergence is visible, albeit slower than linear-TS. When the discretization factor is increased to $M=500$, UCB-1 is unable to explore the entire action space in the $10000$ time steps, and performs very poorly.

In Figure \ref{fig:varyJNR}, we observe the performance of both learning algorithms when the $\jnr$ is variable, and the size of the action space increases as a result. We see that the addition of this selection in the action space renders UCB-1 unable to learn in the limited time horizon examined here. Linear-TS, on the other hand, exhibits learning over the time horizon, although not approaching the optimal strategy within the $10^4$ time steps considered. Thus, we see that while the UCB-1 algorithm is generally unable to learn two continuous parameters simultaneously, the linear-TS approach is capable of learning both $\rho$ and $\jnr$.

In Figure \ref{fig:per}, we examine convergence behavior when $\mathrm{PER}$ is used as a cost function and once again $\jnr$ is selected by the learning algorithm. We observe that for a discretization term of $M = 20$, UCB-1 converges slower than linear-TS, but learning still occurs. When the discretization factor is increased to $M=100$ however, UCB-1 is rendered completely ineffective and the $\mathrm{PER}$ remains constant at zero throughout the simulation.

\section{Conclusion and Open Problems}
\label{se:conclusion}
This paper has presented a linear contextual bandit algorithm that significantly improves finite-time convergence behavior for non-coherent jamming, as compared to the previous state-of-the-art algorithm. This is achieved by introducing context features which exploit inherent similarity between jamming strategies, allowing for a reduced exploration window. Additionally, the use of relevant context features for the non-coherent jamming problem improves performance significantly. Although we have primarily examined the non-coherent case, our results also apply to the coherent setting, and we expect similar improvement in the rate of convergence for both coherent jamming and the case of time offsets examined in \cite{amuru2015optimal,amuru2015jamming}

Future work could focus on a learning setting in which the jammer is able to pay a ``cost" for additional feedback. This is called partial monitoring and has been studied in a general setting \cite[Ch. 37]{Lattimore2020bandit}, but applications to wireless systems are not yet well-explored. Another direction for future work includes study of a learning setting in which the jammer may set less ambitious learning targets in order to improve convergence behavior. This has been analyzed in the context of rate-distortion theory \cite{arumugam2021deciding}, but has not been widely applied to real-world problems. Finally, since knowledge of the optimal jamming strategy for one user may provide relevant information about the optimal jamming strategy for another user, it is possible that online meta-learning and transfer learning techniques can be applied to speed up the learning process by gradually acquiring implicit information about the victims' signals \cite{thornton2022online}.

\bibliographystyle{IEEEtran}
\bibliography{jamBib.bib}{}

\vfill

\end{document}